%% file: main.tex
\documentclass[10pt,twocolumn,letterpaper]{article}
\usepackage[pagenumbers]{cvpr}
\usepackage[letterpaper,margin=1in]{geometry}
\input{preamble}
\input{fp_float_pages}
\input{fp_text_spacing}
\input{fp_vertical_layout}
\title{FP-Prune: Rethinking Visual Token Pruning from an Information Gain Perspective}
\input{authors}
\begin{document}
\flushbottom
\maketitle
\input{sections/abstract}
\input{sections/1_introduction}
% Queue early; fp_float_pages explicitly prevents placement before pages 4--7.
\input{tables/table1}

\input{tables/table2}

\input{tables/table3}

\input{figures/efficiency}
\input{sections/2_related_work}
\input{sections/3_method}

\input{sections/4_experiments}
\input{sections/5_analysis}

\input{sections/6_conclusion}

% The arXiv version lets the disclosure and references follow the main text.
\par
\flushbottom
% Keep the original disclosure before references.
\input{sections/llm_usage}

{\small
\bibliographystyle{ieeenat_fullname}
\bibliography{references}
}

\input{supplementary_body}

\end{document}

%% file: preamble.tex
\usepackage[T1]{fontenc}
\usepackage{iftex}
\ifPDFTeX\usepackage[utf8]{inputenc}\fi
\usepackage{textcomp}
\usepackage{microtype}
\AtBeginDocument{\microtypesetup{expansion=false}}
\usepackage{array}
\usepackage{multirow}
\usepackage{algorithm}
\usepackage{algpseudocode}
\definecolor{cvprblue}{rgb}{0.21,0.49,0.74}
\usepackage[breaklinks,colorlinks,allcolors=cvprblue]{hyperref}
\hypersetup{pdftitle={FP-Prune: Rethinking Visual Token Pruning from an Information Gain Perspective}}
\def\confName{CVPR}
\def\confYear{2027}
\usepackage{float}
\usepackage{colortbl}
\usepackage{fix-cm}
\usepackage[nocheckfootnote]{flushend}

%% file: fp_float_pages.tex
\makeatletter
\newcommand{\FPFloatPage}[1]{%
  \expandafter\xdef\csname FP@minpage@\number\@currbox\endcsname{#1}%
}
\let\FP@original@addtodblcol\@addtodblcol
\def\@addtodblcol{%
  \@ifundefined{FP@minpage@\number\@currbox}{%
    \FP@original@addtodblcol
  }{%
    \ifnum\c@page<\csname FP@minpage@\number\@currbox\endcsname\relax
      \@cons\@deferlist\@currbox
    \else
      \global\expandafter\let\csname FP@minpage@\number\@currbox\endcsname\relax
      \FP@original@addtodblcol
    \fi
  }%
}
\makeatother

%% file: fp_vertical_layout.tex
\makeatletter
\patchcmd{\cvprsection}{-10pt plus -2pt minus -2pt}{-10pt}{}
  {\PackageError{FP-Prune}{Cannot fix section spacing}{Check cvpr.sty}}
\patchcmd{\cvprsubsection}{-8pt plus -2pt minus -2pt}{-8pt}{}
  {\PackageError{FP-Prune}{Cannot fix subsection spacing}{Check cvpr.sty}}
\patchcmd{\cvprsubsubsection}{-6pt plus -2pt minus -2pt}{-6pt}{}
  {\PackageError{FP-Prune}{Cannot fix subsubsection spacing}{Check cvpr.sty}}
\patchcmd{\FPRelatedWorkHeading}{-12pt plus -1pt minus -1pt}{-12pt}{}
  {\PackageError{FP-Prune}{Cannot fix related-work spacing}{Check fp_text_spacing.tex}}

\newcommand{\FPFlexibleLeading}{%
  \ifdim\f@size pt<10.5pt
    \baselineskip=\dimexpr\baselineskip\relax plus .18pt minus .06pt\relax
    \normalbaselineskip=\baselineskip
  \fi
}
\AddToHook{cmd/selectfont/after}{\FPFlexibleLeading}
\AtBeginDocument{\FPFlexibleLeading}
\makeatother

%% file: authors.tex
\author{%
\textbf{Haozhe Zhao}\textsuperscript{1}\thanks{Equal contribution.}\quad
\textbf{Junxiang Zhang}\textsuperscript{1,*}\quad
\textbf{Junxin Wang}\textsuperscript{1,*}\quad
\textbf{Yuan Ge}\textsuperscript{1}\\
\textbf{Xiangchen Song}\textsuperscript{3}\quad
\textbf{Dingyang Zhang}\textsuperscript{1}\quad
\textbf{JingBo Zhu}\textsuperscript{1,2}\quad
\textbf{Tong Xiao}\textsuperscript{1,2,*}\thanks{Corresponding author.}\\[6pt]
\parbox{\dimexpr\textwidth-2\tabcolsep\relax}{\centering\normalsize
\textsuperscript{1}School of Computer Science and Engineering, Northeastern University, Shenyang, China\\
\textsuperscript{2}NiuTrans Research, Shenyang, China\\
\textsuperscript{3}University of Michigan, Ann Arbor, US}%
}
\hypersetup{pdfauthor={Haozhe Zhao, Junxiang Zhang, Junxin Wang, Yuan Ge, Xiangchen Song, Dingyang Zhang, JingBo Zhu, Tong Xiao}}

%% file: sections/abstract.tex
\begin{abstract}
Visual token pruning can reduce the inference cost of vision language models, but its effectiveness depends on preserving complementary visual information under a limited token budget. A central challenge is to quantify the additional information a candidate token contributes beyond the retained set, a contribution that conventional scores based on attention or similarity struggle to capture directly. In this work, we revisit visual token pruning from the perspective of information gain and use mutual information to characterize the incremental value of candidate tokens. We further establish theoretically that frame potential serves as an effective surrogate for mutual information. Building on this observation, we propose \textbf{FP-Prune}, a training-free visual token pruning method that iteratively removes tokens with low information gain to preserve as much complementary visual information as possible. Experiments on three widely used VLMs and eight datasets demonstrate that FP-Prune outperforms baseline methods in both accuracy and inference efficiency. Further analysis shows that the surrogate objective closely aligns with model performance. Our work provides theoretical grounding and an effective \mbox{method for visual token pruning.}
\end{abstract}

%% file: sections/1_introduction.tex
\section{Introduction}\label{introduction}
Vision-Language Models (VLMs) have made significant progress in visual understanding and reasoning tasks, such as visual question answering and image understanding~\cite{ref1,ref2}. These models typically use visual encoders to convert images into sequences of visual tokens, which are then processed by large language models (LLMs). However, visual inputs can introduce a substantial number of additional tokens into the LLM, leading to significant computational and memory costs~\cite{ref3,ref4,ref5}. To address this issue, visual token pruning selectively removes visual tokens while minimizing visual information loss, offering an promising approach \mbox{to efficient VLM inference~\cite{ref5,ref6,ref8}.}

\begin{figure}[t]
  \centering
  % Replace this MMHal-Bench example before release.
  \includegraphics[width=\columnwidth]{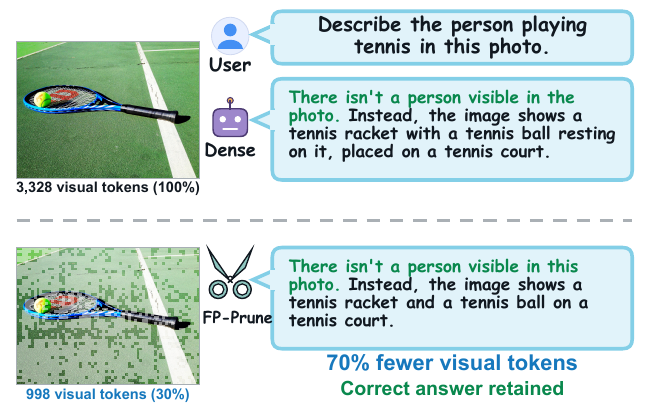}
  \caption{FP-Prune answers the example question correctly while retaining
  30\% of \mbox{the original visual tokens.}}
  \label{fig:fp_intro}
\end{figure}

\begin{figure*}[t]
  \centering
  \includegraphics[width=0.873\textwidth]{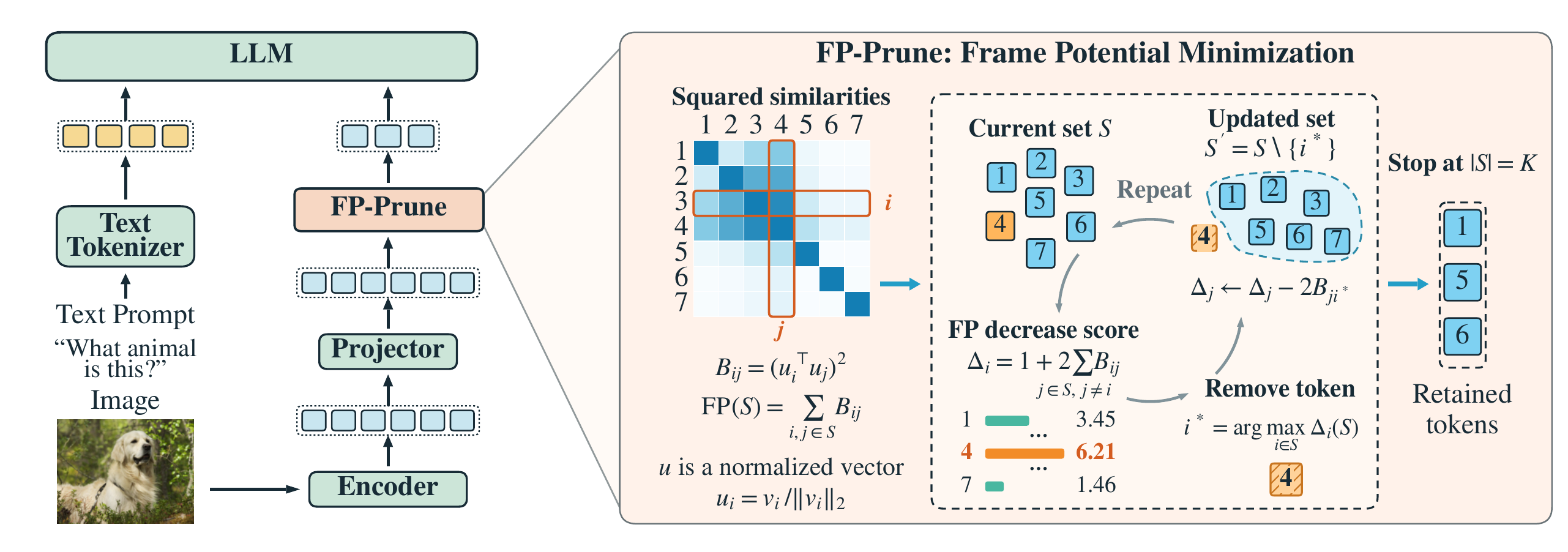}
  \caption{Overview of FP-Prune. The method removes visual tokens in order of
  their current contribution to frame potential, updating the removal scores
  after each step. The retained tokens are passed to the language model \mbox{in their original order.}}
  \label{fig:fp_overview}
\end{figure*}

Existing visual token pruning methods typically select tokens based on attention scores or feature diversity. Attention-based methods leverage internal attention weights as a proxy for token importance, pruning tokens that receive relatively little attention~\cite{ref5,ref6}. In contrast, diversity-based methods focus on feature-space similarity to identify and remove redundant visual tokens~\cite{ref8,ref9}. Although these methods have achieved promising results, their modeling of visual information preservation remains implicit, leading to a gap between their token selection strategies and the goal of maximizing the preservation of original image information. Therefore, an explicit quantitative criterion for measuring information preservation is essential for guiding \mbox{visual token pruning more effectively.}

% Allow natural line breaks and ordinary sentence spacing in this paragraph.
\begingroup\microtypesetup{expansion=true}\frenchspacing
\hfuzz=0pt\hbadness=1000
To this end, we revisit visual token pruning from the perspective of information gain. A token's information contribution depends not only on its own information content but also on how much additional information it provides to the retained set. Mutual information measures the information shared between tokens~\cite{cover2006elements}. When a token's information is already represented by others, it contributes little new information, whereas tokens that complement existing representations provide greater value. Based on this observation, we use pairwise mutual information to characterize redundancy among retained tokens. Through a second-order approximation, we establish a mathematical connection between pairwise mutual information and Frame Potential and derive an explicit lower bound on retained information. We further prove that, under a fixed token budget, minimizing Frame Potential is equivalent to maximizing \mbox{this lower bound.}\par\endgroup

Based on the theoretical analysis, we propose FP-Prune, a training-free visual token pruning method. FP-Prune uses Frame Potential to measure information redundancy within the retained token set. By iteratively removing the token that yields the largest reduction in Frame Potential, FP-Prune aims to preserve as much complementary visual information as possible under a given token budget. We conduct extensive experiments on three widely used VLMs and eight datasets. The results demonstrate that FP-Prune outperforms baseline methods in both task performance and inference efficiency, with particularly strong performance on text-related and \mbox{fine-grained visual understanding tasks.}\looseness=1\par

Our contributions are threefold. (1) We link Frame Potential to pairwise
information redundancy and show that minimizing it maximizes a lower bound
on retained information at a fixed token budget. (2) We introduce FP-Prune,
a training-free pruning method with greedy token removal and incremental
score updates. (3) Across three VLMs and eight benchmarks, FP-Prune
outperforms the evaluated pruning baselines in \mbox{task performance and inference efficiency.}

%% file: tables/table1.tex
\begin{table*}[t]
\FPFloatPage{4}

\centering

\caption{Task performance of LLaVA-NeXT-Vicuna-7B at three visual token retention budgets.}

\label{tab:fp_llava_next}

\begingroup

\fontsize{9.5}{11.4}\selectfont

\setlength{\tabcolsep}{3pt}

\renewcommand{\arraystretch}{1.18}

% Reference styling: gray Ret column and FP-Prune rows; full-width budget headings.

\arrayrulecolor{black}

\setlength{\arrayrulewidth}{0.45pt}

\setlength{\aboverulesep}{0pt}

\setlength{\belowrulesep}{0pt}

\setlength{\abovetopsep}{0pt}

\setlength{\belowbottomsep}{0pt}

\resizebox{0.98\textwidth}{!}{%

\begin{tabular}{p{38.8pt}*{8}{>{\centering\arraybackslash}p{38.8pt}}|>{\centering\arraybackslash}p{38.8pt}|>{\columncolor[gray]{0.96}\centering\arraybackslash}p{38.8pt}}

\toprule

\textbf{Method} & \textbf{GQA} & \textbf{TextVQA} & \textbf{MMB} & \textbf{SQA} & \textbf{OCR} & \textbf{MME} & \textbf{MMMU} & \textbf{MMStar} & \textbf{Avg} & \cellcolor[gray]{0.87}\textbf{Ret (\%)} \\

\midrule

Dense & 64.22 & 64.88 & 67.10 & 70.10 & 52.10 & 1844.06 & 36.56 & 37.48 & 57.29 & 100.00 \\

\midrule

\multicolumn{11}{c}{\cellcolor[gray]{0.96}\rule{0pt}{2.4ex}\textbf{Visual token retention}\quad 10\%} \\

DivPrune & 59.22 & 47.86 & 61.68 & 67.38 & 27.80 & 1615.90 & \textbf{36.89} & 33.45 & 49.00 & 85.65 \\

MMTok & \textbf{60.26} & 52.70 & 62.80 & \textbf{68.27} & 34.60 & \textbf{1726.26} & 35.67 & 33.61 & 51.20 & 89.16 \\

OTPrune & 59.69 & 49.93 & 62.54 & \textbf{68.27} & 30.80 & 1676.73 & 36.00 & 34.24 & 50.17 & 87.55 \\

\rowcolor[gray]{0.93}

FP-Prune & 59.59 & \textbf{58.49} & \textbf{64.35} & 67.33 & \textbf{36.00} & 1718.57 & 35.33 & \textbf{34.35} & \textbf{52.10} & \cellcolor[gray]{0.85}\textbf{90.69} \\

\midrule

\multicolumn{11}{c}{\cellcolor[gray]{0.96}\rule{0pt}{2.4ex}\textbf{Visual token retention}\quad 20\%} \\

DivPrune & 62.17 & 53.10 & 64.09 & 67.53 & 33.50 & 1717.73 & 37.44 & 35.15 & 51.79 & 90.52 \\

MMTok & \textbf{62.79} & 57.23 & 64.52 & \textbf{67.58} & 39.20 & 1764.96 & \textbf{37.78} & \textbf{36.12} & 53.53 & 93.65 \\

OTPrune & 62.11 & 57.68 & 65.21 & 67.28 & 40.00 & 1767.48 & \textbf{37.78} & 35.61 & 53.60 & 93.72 \\

\rowcolor[gray]{0.93}

FP-Prune & 61.83 & \textbf{61.87} & \textbf{65.72} & 67.13 & \textbf{41.50} & \textbf{1782.91} & 35.78 & 35.81 & \textbf{54.16} & \cellcolor[gray]{0.85}\textbf{94.39} \\

\midrule

\multicolumn{11}{c}{\cellcolor[gray]{0.96}\rule{0pt}{2.4ex}\textbf{Visual token retention}\quad 30\%} \\

DivPrune & \textbf{63.37} & 56.57 & 64.86 & \textbf{68.22} & 38.30 & 1749.72 & \textbf{38.11} & 35.74 & 53.46 & 93.48 \\

MMTok & 63.36 & 60.11 & 65.29 & 68.02 & 43.70 & 1795.77 & 37.33 & 35.86 & 54.73 & 95.59 \\

OTPrune & 63.15 & 61.16 & \textbf{65.98} & 67.92 & 44.00 & \textbf{1807.42} & 36.67 & \textbf{35.97} & 54.92 & 95.82 \\

\rowcolor[gray]{0.93}

FP-Prune & 63.04 & \textbf{63.14} & 65.89 & 67.72 & \textbf{47.40} & 1800.83 & 36.67 & 35.91 & \textbf{55.51} & \cellcolor[gray]{0.85}\textbf{96.88} \\

\bottomrule

\end{tabular}

}

\endgroup

\end{table*}

%% file: tables/table2.tex
\begin{table*}[t]
\FPFloatPage{5}

\centering

\caption{Task performance of Qwen2.5-VL-7B at three visual token retention budgets.}

\label{tab:fp_qwen}

\begingroup

\fontsize{9.5}{11.4}\selectfont

\setlength{\tabcolsep}{3pt}

\renewcommand{\arraystretch}{1.18}

% Reference styling: gray Ret column and FP-Prune rows; full-width budget headings.

\arrayrulecolor{black}

\setlength{\arrayrulewidth}{0.45pt}

\setlength{\aboverulesep}{0pt}

\setlength{\belowrulesep}{0pt}

\setlength{\abovetopsep}{0pt}

\setlength{\belowbottomsep}{0pt}

\resizebox{0.98\textwidth}{!}{%

\begin{tabular}{p{38.8pt}*{8}{>{\centering\arraybackslash}p{38.8pt}}|>{\centering\arraybackslash}p{38.8pt}|>{\columncolor[gray]{0.96}\centering\arraybackslash}p{38.8pt}}

\toprule

\textbf{Method} & \textbf{GQA} & \textbf{TextVQA} & \textbf{MMB} & \textbf{SQA} & \textbf{OCR} & \textbf{MME} & \textbf{MMMU} & \textbf{MMStar} & \textbf{Avg} & \cellcolor[gray]{0.87}\textbf{Ret (\%)} \\

\midrule

Dense & 61.04 & 83.88 & 83.33 & 88.05 & 85.70 & 2315.31 & 50.22 & 62.67 & 74.70 & 100.00 \\

\midrule

\multicolumn{11}{c}{\cellcolor[gray]{0.96}\rule{0pt}{2.4ex}\textbf{Visual token retention}\quad 10\%} \\

DivPrune & 56.12 & 63.22 & 76.03 & 81.36 & 48.10 & 2030.69 & \textbf{46.56} & 47.38 & 61.41 & 82.89 \\

MMTok & 55.13 & 59.48 & 75.17 & 80.81 & 44.10 & 2015.07 & 44.44 & 46.57 & 59.71 & 80.56 \\

OTPrune & \textbf{57.64} & 39.08 & 75.69 & \textbf{83.24} & 33.60 & 1951.95 & 45.11 & 47.02 & 56.39 & 76.84 \\

\rowcolor[gray]{0.93}

FP-Prune & 55.05 & \textbf{69.06} & \textbf{77.23} & 83.09 & \textbf{50.80} & \textbf{2127.79} & 45.89 & \textbf{48.07} & \textbf{63.15} & \cellcolor[gray]{0.85}\textbf{84.85} \\

\midrule

\multicolumn{11}{c}{\cellcolor[gray]{0.96}\rule{0pt}{2.4ex}\textbf{Visual token retention}\quad 20\%} \\

DivPrune & 58.79 & 74.05 & \textbf{79.64} & 84.38 & \textbf{64.20} & 2184.20 & 47.22 & \textbf{54.00} & 67.54 & 90.68 \\

MMTok & 58.14 & 71.81 & 79.38 & 83.74 & 61.10 & 2209.18 & \textbf{48.00} & 53.36 & 66.80 & 89.83 \\

OTPrune & \textbf{59.45} & 51.77 & 78.09 & 84.28 & 46.30 & 2067.02 & 46.33 & 52.86 & 61.61 & 83.56 \\

\rowcolor[gray]{0.93}

FP-Prune & 57.76 & \textbf{78.08} & 79.21 & \textbf{84.73} & 61.80 & \textbf{2224.49} & 47.67 & 53.61 & \textbf{67.79} & \cellcolor[gray]{0.85}\textbf{90.96} \\

\midrule

\multicolumn{11}{c}{\cellcolor[gray]{0.96}\rule{0pt}{2.4ex}\textbf{Visual token retention}\quad 30\%} \\

DivPrune & 59.76 & 78.59 & 80.58 & 85.37 & \textbf{72.90} & 2206.26 & 46.67 & 56.79 & 69.93 & 93.64 \\

MMTok & 59.36 & 76.73 & \textbf{81.79} & 85.13 & 70.20 & 2268.01 & 48.67 & \textbf{56.87} & 69.97 & 93.88 \\

OTPrune & \textbf{59.88} & 61.27 & 79.98 & 84.38 & 59.20 & 2205.73 & 47.67 & 55.63 & 65.85 & 88.87 \\

\rowcolor[gray]{0.93}

FP-Prune & 58.56 & \textbf{81.77} & 81.36 & \textbf{86.66} & 69.40 & \textbf{2286.06} & \textbf{49.44} & 56.64 & \textbf{70.69} & \cellcolor[gray]{0.85}\textbf{94.76} \\

\bottomrule

\end{tabular}

}

\endgroup

\end{table*}

%% file: tables/table3.tex
\begin{table*}[t]
\FPFloatPage{6}

\centering

\caption{Task performance of InternVL3.5-8B at three visual token retention budgets.}

\label{tab:fp_internvl}

\begingroup

\fontsize{9.5}{11.4}\selectfont

\setlength{\tabcolsep}{3pt}

\renewcommand{\arraystretch}{1.18}

% Reference styling: gray Ret column and FP-Prune rows; full-width budget headings.

\arrayrulecolor{black}

\setlength{\arrayrulewidth}{0.45pt}

\setlength{\aboverulesep}{0pt}

\setlength{\belowrulesep}{0pt}

\setlength{\abovetopsep}{0pt}

\setlength{\belowbottomsep}{0pt}

\resizebox{0.98\textwidth}{!}{%

\begin{tabular}{p{38.8pt}*{8}{>{\centering\arraybackslash}p{38.8pt}}|>{\centering\arraybackslash}p{38.8pt}|>{\columncolor[gray]{0.96}\centering\arraybackslash}p{38.8pt}}

\toprule

\textbf{Method} & \textbf{GQA} & \textbf{TextVQA} & \textbf{MMB} & \textbf{SQA} & \textbf{OCR} & \textbf{MME} & \textbf{MMMU} & \textbf{MMStar} & \textbf{Avg} & \cellcolor[gray]{0.87}\textbf{Ret (\%)} \\

\midrule

Dense & 61.62 & 78.11 & 83.42 & 97.32 & 82.40 & 2363.55 & 57.78 & 66.27 & 76.42 & 100.00 \\

\midrule

\multicolumn{11}{c}{\cellcolor[gray]{0.96}\rule{0pt}{2.4ex}\textbf{Visual token retention}\quad 10\%} \\

DivPrune & 56.50 & 52.95 & 74.83 & 85.18 & 29.40 & 2004.98 & 51.11 & 51.36 & 59.12 & 77.90 \\

MMTok & \textbf{58.87} & 34.18 & 73.71 & 85.92 & 13.50 & 2003.42 & 50.56 & 48.72 & 54.63 & 72.26 \\

OTPrune & 57.89 & 21.64 & 73.71 & 87.51 & 9.10 & 1960.48 & 51.00 & 51.38 & 52.78 & 69.97 \\

\rowcolor[gray]{0.93}

FP-Prune & 56.50 & \textbf{61.43} & \textbf{75.69} & \textbf{87.65} & \textbf{42.10} & \textbf{2018.58} & \textbf{53.56} & \textbf{52.41} & \textbf{62.68} & \cellcolor[gray]{0.85}\textbf{82.43} \\

\midrule

\multicolumn{11}{c}{\cellcolor[gray]{0.96}\rule{0pt}{2.4ex}\textbf{Visual token retention}\quad 20\%} \\

DivPrune & 58.94 & 63.81 & 78.44 & 90.88 & 44.40 & 2129.42 & 53.78 & 57.24 & 65.44 & 86.02 \\

MMTok & \textbf{60.58} & 44.73 & 77.92 & 90.63 & 26.80 & 2101.97 & 52.00 & 55.77 & 60.44 & 79.71 \\

OTPrune & 59.76 & 29.95 & 77.84 & 89.69 & 21.90 & 2087.92 & 52.56 & 56.03 & 57.79 & 76.40 \\

\rowcolor[gray]{0.93}

FP-Prune & 59.23 & \textbf{70.71} & \textbf{80.07} & \textbf{93.55} & \textbf{58.80} & \textbf{2173.93} & \textbf{54.78} & \textbf{57.33} & \textbf{69.01} & \cellcolor[gray]{0.85}\textbf{90.43} \\

\midrule

\multicolumn{11}{c}{\cellcolor[gray]{0.96}\rule{0pt}{2.4ex}\textbf{Visual token retention}\quad 30\%} \\

DivPrune & 59.50 & 69.72 & 80.15 & 93.75 & 56.90 & 2202.14 & 55.00 & 60.16 & 69.23 & 90.80 \\

MMTok & \textbf{60.90} & 51.67 & 80.15 & 92.46 & 42.00 & 2185.89 & 53.22 & 59.52 & 64.75 & 85.18 \\

OTPrune & 60.47 & 36.51 & 79.47 & 92.51 & 36.10 & 2176.71 & 53.89 & 58.80 & 61.94 & 81.64 \\

\rowcolor[gray]{0.93}

FP-Prune & 60.49 & \textbf{74.72} & \textbf{82.04} & \textbf{94.89} & \textbf{68.20} & \textbf{2283.62} & \textbf{56.67} & \textbf{61.13} & \textbf{72.46} & \cellcolor[gray]{0.85}\textbf{94.92} \\

\bottomrule

\end{tabular}

}

\endgroup

\end{table*}

%% file: figures/efficiency.tex
\begin{figure*}[t]
\FPFloatPage{7}
  \centering
  \captionsetup{skip=2bp}
  \includegraphics[width=\textwidth]{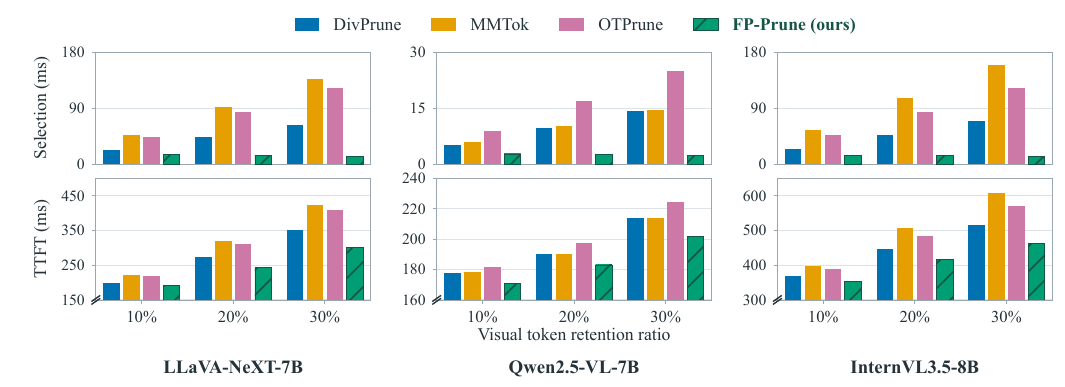}
  \caption{Inference efficiency at different visual token retention budgets. Each model panel shows token selection latency above TTFT, with a shared horizontal axis. Both metrics are in milliseconds (lower is better). Selection latency axes start at zero. The break marks indicate that TTFT axes start above zero to show differences among methods. Green hatched bars denote FP-Prune. Benchmark means are weighted by \mbox{the full question counts.}}
  \label{fig:fp_efficiency}
\end{figure*}

%% file: sections/2_related_work.tex
\FPRelatedWorkHeading

\subsection{Vision-Language Models}

Vision-language models connect visual encoders with large language models through cross-modal connectors, integrating visual perception with language understanding and generation. LLaVA supports image-based question answering and interaction through visual instruction tuning~\cite{ref1}, while BLIP-2 connects vision and language models through image-text pretraining~\cite{ref2}. As applications expand to text recognition, document parsing, and fine-grained understanding, models further enhance their ability to process high-resolution visual inputs. Qwen2.5-VL uses dynamic resolution to accommodate images of different sizes~\cite{ref3}; InternVL3.5 preserves local details in high-resolution images through dynamic tiling~\cite{ref4}. These mechanisms provide richer visual information but also produce longer visual token sequences, increasing the computational and memory overhead of language models. Therefore, reducing token counts while preserving useful visual information has become an important research direction \mbox{for improving inference efficiency~\cite{ref5,ref6,ref8}.}

\subsection{Visual Token Pruning}

\paragraph{Attention-based.}
Attention-based methods use attention weights as a proxy for visual token importance. FastV~\cite{ref5} uses attention scores from early layers to prune visual tokens for subsequent computation. SparseVLM~\cite{ref6} selects relevant text tokens to assess visual token significance through attention and progressively sparsifies the visual sequence. Beyond attention alone, TransPrune~\cite{ref7} combines changes in token representations across layers with instruction-guided attention. However, selecting highly attended tokens can miss complementary visual context, since attention-based saliency alone does not \mbox{ensure broad semantic coverage~\cite{ref10}.}

\paragraph{Diversity-based.}
Diversity-based methods exploit feature relationships to reduce duplication and preserve complementary visual content~\cite{ref8,ref9}. Related work further considers subset coverage, sometimes jointly with saliency or prompt alignment, to improve the representativeness of retained tokens~\cite{ref10,ref11,ref12}. For example, DivPrune~\cite{ref8} maximizes the minimum pairwise distance among selected tokens. MMTok~\cite{dong2026mmtok} selects visual tokens to jointly cover the original visual tokens and text tokens, while OTPrune~\cite{fp_chen2026otprune} aligns the full and retained visual token distributions through optimal transport. Our analysis shows that minimizing Frame Potential at a fixed token budget maximizes a lower bound on retained information, motivating its use as a surrogate objective to \mbox{guide visual token pruning.}\looseness=1\par

%% file: sections/3_method.tex
\section{Method}\label{sec:method}
\subsection{Problem Definition}\label{sec:formulation}

Given an input image $x$, a visual encoder converts it into $N$ visual tokens, $V=[v_1,\ldots,v_N]\in\mathbb{R}^{N\times d}$, where $v_i\in\mathbb{R}^{d}$ and $d$ is the feature dimension. Visual token pruning selects a subset of $K$ tokens, $V_S=\{v_i\mid i\in S\}$, with $S\subseteq\{1,\ldots,N\}$ and $|S|=K$. The goal is to reduce the number of visual tokens while retaining as much information from the original representation as possible. \mbox{We formulate this goal as}
\begin{equation}
S^*=\arg\max_{\substack{S\subseteq\{1,\ldots,N\}\\|S|=K}}\mathcal{R}(V_S;V),
\label{eq:retention_objective}
\end{equation}
where $\mathcal{R}(V_S;V)$ quantifies the information preserved by the selected subset relative to the full visual representation. Importantly, the objective concerns the \emph{retained set as a whole}, rather than the importance of each token considered independently: selecting individually informative tokens need not preserve more information if \mbox{they encode largely overlapping content.}

\subsection{Mutual Information-Based Formulation}

Following prior work on information-theoretic subset selection
\cite{guestrin2005near,Das_2025_CVPR},
we adopt mutual information to formulate the information-retention objective.
In particular, mutual information quantifies how much information
the selected tokens retain about the latent information source associated with the image. Thus, we express the information-retention objective in terms \mbox{of mutual information as:}
\begin{equation}
\begin{aligned}
S^* &= \arg\max_{\substack{S\subseteq\{1,\ldots,N\}\\|S|=K}}\mathcal{R}(V_S;V) \\
&= \arg\max_{|S|=K} I(Z;T_S),
\end{aligned}
\end{equation}
where $I(\cdot;\cdot)$ denotes mutual information,
$Z$ is a latent random variable representing the
latent information source associated with the original visual representation,
$T_i$ denotes the random representation of visual token $i$,
and $T_S$ denotes the joint representation \mbox{of the selected tokens.}

For an arbitrary retained set $S=\{i_1,\ldots,i_K\}$, let $S_k=\{i_1,\ldots,i_k\}$, where $S_0=\varnothing$. According to the chain rule of mutual information~\cite{cover2006elements}, the overall information contained in the retained set can be decomposed into the sum of the information gains contributed \mbox{by sequentially adding each token:}
\begin{equation}
\begin{aligned}
I(Z;T_S)
&=\sum_{k=1}^{K}
\left[
I(Z;T_{S_k})-I(Z;T_{S_{k-1}})
\right]\\
&=\sum_{k=1}^{K}
I(Z;T_{i_k}\mid T_{S_{k-1}}).
\end{aligned}
\end{equation}

Accordingly, the information gain of a candidate token $i\notin S$ with respect to the current retained \mbox{set $S$ can be expressed as:}
\begin{equation}
I(Z;T_i\mid T_S)
=
I(Z;T_{S\cup\{i\}})-I(Z;T_S).
\end{equation}

Therefore, for a given retained set, maximizing the information gain of a candidate token is equivalent to maximizing the change in the information content of the set before and after adding that token. However, computing both mutual information terms requires the joint probability distribution of high-dimensional visual representations, making their direct quantification challenging. Therefore, \mbox{to guide visual token pruning,} it is necessary to establish a computable criterion that has a clear theoretical connection to the information retention objective, and captures the additional information a candidate token provides beyond what is already \mbox{represented by the retained set.}

\subsection{Frame Potential as an Information Proxy}
\label{sec:fp_proxy}

To analyze the dependence among selected tokens in a tractable probabilistic setting, we adopt a linear
Gaussian observation model that links visual token features
to their associated random representations.
Given a visual token feature $v_i$ produced by the visual
encoder, we normalize it as $u_i=v_i/\|v_i\|_2$,
and let the shared latent information source be
$Z\sim\mathcal{N}(0,I_d)$.
The random representation corresponding to the $i$-th
\mbox{token is defined as}
\begin{equation}
T_i=u_i^\top Z+\varepsilon_i,
\label{eq:gaussian_model}
\end{equation}
where $\varepsilon_i\sim\mathcal{N}(0,\sigma^2)$ denotes
Gaussian noise that is mutually independent across tokens
and independent of $Z$, with $\sigma^2>0$ being \mbox{the
shared noise variance.}

For a retained set $S$, we stack the corresponding random
representations into a vector $T_S$ and denote the matrix
of normalized token features \mbox{by $U_S\in\mathbb{R}^{K\times d}$. Then, we have}
\begin{equation}
T_S=U_SZ+\varepsilon_S,\qquad
\varepsilon_S\sim\mathcal{N}(0,\sigma^2 I_K),
\end{equation}
where $I_K$ denotes the $K$-dimensional identity matrix.
Accordingly, the information retained by the selected
set can be expressed \mbox{in closed form as:}
\begin{equation}
\label{eq:logdet_mi}
I(Z;T_S)=\frac{1}{2}\log\det\left(
I_K+\frac{1}{\sigma^2}U_SU_S^\top
\right).
\end{equation}
The full derivation of this result can be found in Sec.~B.1 \mbox{of the supplementary material.}

\begingroup\microtypesetup{expansion=true}\frenchspacing
\hfuzz=0pt\hbadness=1000
\paragraph{Total Correlation and Frame Potential.}
Although Eq.~\eqref{eq:logdet_mi} provides a closed-form
expression for a given retained set, using it directly
during iterative pruning requires repeated log-determinant
evaluations over candidate subsets.
To obtain a more tractable feature-based criterion,
we introduce total correlation (TC)~\cite{watanabe1960information}
to characterize the statistical dependence \mbox{among selected tokens:}
% To characterize the statistical dependence among selected
% tokens, we introduce total correlation (TC)~\cite{watanabe1960information},
% which measures the deviation of their joint distribution
% from the product of their marginal distributions:
\begin{equation}
\operatorname{TC}(T_S)
=
D_{\mathrm{KL}}\left(
p(T_S)\middle\|
\prod_{i\in S}p(T_i)
\right).
\end{equation}
\par\endgroup

% Under our Gaussian observation model, the token
% representations are conditionally independent given $Z$.
% Moreover, since all token features are L2-normalized
% and share the same noise variance, their individual
% mutual information with $Z$ is identical.
% Consequently, as derived in Supplementary Sec.~B.2,
% the information retained by the selected set can be
% expressed as
% \begin{equation}
% I(Z;T_S)
% =
% \frac{K}{2}\log\left(1+\frac{1}{\sigma^2}\right)
% -\operatorname{TC}(T_S).
% \label{eq:mi_tc}
% \end{equation}
% Thus, for a fixed token budget and noise variance,
% maximizing the retained mutual information is equivalent
% to minimizing the total correlation among selected tokens.
Under our Gaussian observation model, the token
representations are conditionally independent given $Z$.
Moreover, since all token features are L2-normalized
and share the same noise variance, their individual
mutual information with $Z$ is identical.
By the definition of total correlation and the conditional
independence among token representations, the information
retained by the selected set \mbox{can be expressed as}\looseness=1
\begin{equation}
I(Z;T_S)
=
\frac{K}{2}\log\left(1+\frac{1}{\sigma^2}\right)
-\operatorname{TC}(T_S).
\label{eq:mi_tc}
\end{equation}
The derivation is provided in Supplementary Sec.~B.2.
Thus, for a fixed token budget and noise variance,
maximizing the retained mutual information is equivalent
to minimizing the total \mbox{correlation among selected tokens.}

% To obtain a feature-based measure of this statistical
% dependence, we connect total correlation to the Frame
% Potential~\cite{benedetto2003finite}.
To obtain a tractable approximation for minimizing total correlation,
we analyze its second-order expansion under the Gaussian observation model.
Specifically, let $G_S=U_SU_S^\top$ denote the Gram matrix of the
normalized token features, and define the Frame Potential~\cite{benedetto2003finite} as
$\mathrm{FP}(S)=\operatorname{tr}(G_S^2)
=\sum_{i,j\in S}(u_i^\top u_j)^2$.
Under the Gaussian observation model, a second-order
Taylor expansion of the total correlation around
\mbox{zero pairwise correlations yields}
\begin{equation}
\operatorname{TC}(T_S)
\approx
\frac{\mathrm{FP}(S)-K}{4(1+\sigma^2)^2}.
\label{eq:tc_fp}
\end{equation}
The complete derivation of Eq.~\eqref{eq:tc_fp} is \mbox{provided in Supplementary Sec.~B.2.}

This relationship establishes Frame Potential as a
second-order approximation to the total correlation
among selected tokens. Combined with Eq.~\eqref{eq:mi_tc},
it provides a direct connection between minimizing
Frame Potential and maximizing the information retained
by the selected set \mbox{under the weak-correlation approximation.}

\paragraph{A Lower Bound on Retained Information.}
The preceding result connects Frame Potential to
retained mutual information through a weak-correlation
approximation. To obtain a complementary guarantee
that does not rely on this approximation, we derive
a lower bound directly from the log-determinant
expression in Eq.~\eqref{eq:logdet_mi}.
By applying eigendecomposition to the Gram matrix and
a second-order lower bound to the resulting \mbox{logarithmic
terms, we obtain:}

\begin{equation}
I(Z;T_S)\geq
\frac{K}{2\sigma^2}
-\frac{\mathrm{FP}(S)}{4\sigma^4}.
\label{eq:mi_lower_bound}
\end{equation}
The corresponding proof is provided in Supplementary Sec.~B.3.
Unlike the preceding approximation, this lower
bound does not rely on a weak-correlation assumption
and holds for any $\sigma^2>0$ under the Gaussian
observation model. For a fixed token budget $K$ and noise variance
$\sigma^2$, minimizing Frame Potential is equivalent
to maximizing this lower bound.
Therefore, Frame Potential provides a theoretically
grounded and directly computable criterion \mbox{for
visual token pruning.} Its pairwise additive structure further enables efficient greedy pruning by allowing token removal scores \mbox{to be updated incrementally.}

\subsection{FP-Prune}\label{sec:fp_prune}

% Queue the pseudocode at the start of Section 3.4 so it can sit beside its prose.
\begin{algorithm}[!t]
\caption{FP-Prune}
\label{alg:fp_prune}
\begin{algorithmic}[1]
\Require Visual tokens $V=[v_1,\ldots,v_N]$; budget $K$
\Ensure $K$ retained visual tokens in their original order
\State $u_i\gets v_i/\|v_i\|_2$ for $i=1,\ldots,N$
\State $A_{ij}\gets (u_i^\top u_j)^2$ for all $i,j$
\State $S_N\gets\{1,\ldots,N\}$
\State $\Delta_i(S_N)\gets 2\sum_{j=1}^{N} A_{ij}-1$ for all $i$
\For{$k=N,N-1,\ldots,K+1$}
  \State $i^*\gets\arg\max_{i\in S_k}\Delta_i(S_k)$
  \State Form $S_{k-1}$ by removing $i^*$ from $S_k$
  \ForAll{$j\in S_{k-1}$}
    \State $\Delta_j(S_{k-1})\gets\Delta_j(S_k)-2A_{ji^*}$
  \EndFor
\EndFor
\State Sort the indices in $S_K$ in ascending order
\State \Return the corresponding original tokens from $V$
\end{algorithmic}
\end{algorithm}

% Natural spacing for the two introductory FP-Prune paragraphs.
\begingroup\microtypesetup{expansion=true}\frenchspacing
\hfuzz=0pt\hbadness=1000
% Slightly more open word spacing, confined to this paragraph.
\begingroup\spaceskip=2.7pt plus .7pt minus .3pt\relax
At a fixed token budget $K$, minimizing Frame Potential maximizes the
lower bound on the retained information in Eq.~\eqref{eq:mi_lower_bound}.
FP-Prune uses this criterion to select visual tokens without any additional
training. It starts with the full set of visual tokens and greedily removes one
token at a time until $K$ tokens remain~\cite{ranieri2014near}. Starting from the full set allows the
redundancy of each token to be assessed through its interactions with all other
tokens before any token is discarded. It also avoids an arbitrary choice of the initial
token when building a set from scratch: after normalization, every
singleton has Frame Potential equal to one, so the objective alone cannot
distinguish the first \mbox{token to add.}\par\endgroup

Let $S_k$ denote the current set of $k$ retained visual tokens. The removal
score $\Delta_i(S_k)$ measures how much the Frame Potential decreases when token $i$ is removed \mbox{from this set:}
\begin{equation}
\FPCompactDisplayBelow
\Delta_i(S_k)=1+2\sum_{\substack{j\in S_k\\j\neq i}}(u_i^\top u_j)^2.
\label{eq:removal_score}
\end{equation}
\par\endgroup
At each iteration, FP-Prune removes the token $i^*$ with the largest
removal score in the $S_k$. The remaining tokens form $S_{k-1}$, and the
score of each remaining token $j$ is updated by subtracting its
interaction with \mbox{the removed token:}
\begin{equation}
\Delta_j(S_{k-1})
=\Delta_j(S_k)-2(u_j^\top u_{i^*})^2.
\label{eq:incremental_update}
\end{equation}
This update avoids recomputing the scores from scratch. All pairwise similarities are computed once before pruning and then reused for score updates throughout the deletion process. At budget $K$,
FP-Prune restores the original order of the tokens in $S_K$ by sorting
them by their original indices before passing them to the language model
for inference. \mbox{Algorithm~\ref{alg:fp_prune} summarizes the procedure.}

%% file: sections/4_experiments.tex
% Experimental prose; table definitions are maintained in tables/fp_*.tex.
% Optional legacy float placement; leave disabled while drafting.
% \input{fp_prune_layout}

\section{Experiments}
\label{sec:fp_experiments}

\subsection{Experimental Setup}
\label{sec:fp_setup}

\noindent\textbf{Models.}
We evaluate FP-Prune on LLaVA-NeXT-Vicuna-7B~\cite{liu2024llavanext}, Qwen2.5-VL-7B~\cite{ref3}, and InternVL3.5-8B~\cite{ref4}. These models construct visual token sequences using different strategies: resolution-dependent cropping in LLaVA-NeXT, native dynamic resolution in Qwen2.5-VL, \mbox{and dynamic tiling in InternVL3.5.}

\noindent\textbf{Evaluation.}
We use eight benchmarks: GQA~\cite{hudson2019gqa}, TextVQA~\cite{singh2019textvqa}, MMBench-EN~\cite{liu2024mmbench}, ScienceQA-IMG~\cite{lu2022scienceqa}, OCRBench~\cite{liu2024ocrbench}, MME~\cite{fu2023mme}, MMMU~\cite{yue2024mmmu}, and MMStar~\cite{chen2024mmstar}.
Avg is the unweighted mean of the eight scores on a 0--100 scale, with MME rescaled from its maximum of 2,800. Ret (\%) is the unweighted mean of the eight per-benchmark score ratios to the corresponding Dense model, \mbox{expressed as a percentage.}

\smallskip
\noindent We evaluate efficiency using token selection latency, TTFT, and cumulative end-to-end GPU time. Selection latency measures visual token selection time, and TTFT measures the delay to the first output token. Each FP-Prune or Dense configuration covers eight full datasets and 29,698 questions. Cumulative GPU time sums each GPU's effective time per dataset, then across all eight datasets, excluding model loading, warm-up, and synchronization waits. End-to-end speedup is the Dense/FP-Prune ratio \mbox{of cumulative GPU times.}

\smallskip
\noindent\textbf{Baselines.}
We compare FP-Prune with three training-free baselines. DivPrune~\cite{ref8} selects visual tokens based on diversity, MMTok~\cite{dong2026mmtok} maximizes multimodal coverage, and OTPrune~\cite{fp_chen2026otprune} aligns token distributions. Dense retains all visual tokens within the model's context limit and serves as the reference \mbox{for Ret and speedup.}

\smallskip
\noindent\textbf{Implementation details.}
FP-Prune selects visual tokens before they enter the language model and requires no additional training. For each model, we evaluate all pruning methods at per-image retention rates of 10\%, 20\%, and 30\% with identical inputs, prompts, and scoring procedures. All methods follow the model's native image preprocessing pipeline and use greedy decoding. Token selection latency and TTFT are measured on a single NVIDIA RTX 3090 GPU with 24\,GB of \mbox{memory at batch size 1.}

\subsection{Main Results}
\label{sec:fp_results}

As shown in Tables~\ref{tab:fp_llava_next}--\ref{tab:fp_internvl}, FP-Prune achieves the highest Avg and Ret among the pruning methods across all nine configurations spanning three models and three visual token retention budgets. With only 30\% of the visual tokens retained, it preserves 96.88\%, 94.76\%, and 94.92\% of the Dense model's performance on LLaVA-NeXT, Qwen, and InternVL, respectively. These results demonstrate consistent performance advantages across models \mbox{with different visual tokenization strategies.}

The advantage is most pronounced at the lowest retention budget on LLaVA-NeXT and Qwen. With 10\% of the visual tokens retained, FP-Prune improves Avg over the strongest baseline on each model by 0.91 points on LLaVA-NeXT, 1.74 points on Qwen, and 3.56 points on InternVL. The corresponding baselines are MMTok for LLaVA-NeXT and DivPrune for Qwen and InternVL. At higher budgets, FP-Prune remains ahead in every configuration, with gains of 3.57 and 3.23 points over DivPrune on InternVL \mbox{at 20\% and 30\% retention.}

FP-Prune provides particularly strong gains on text-related tasks, achieving the best TextVQA score among the pruning methods in all nine configurations. At 10\% retention, it exceeds the strongest TextVQA baseline by 5.79, 5.84, and 8.48 percentage points on LLaVA-NeXT, Qwen, and InternVL, respectively. On InternVL, it also outperforms DivPrune, the strongest OCRBench baseline at each budget, by 11.30--14.40 points across the three budgets. These results suggest that FP-Prune better preserves performance on text-related tasks when the \mbox{visual token budget is limited.}

\subsection{Inference Efficiency}
\label{sec:fp_efficiency}

As shown in Figure~\ref{fig:fp_efficiency}, FP-Prune achieves the lowest token selection latency and TTFT among pruning methods on all three models at every retention budget. Latencies are averaged over the eight benchmarks, weighted by their \mbox{full evaluation question counts.}

% Use the same column height on both sides of the arXiv page.
\input{figures/analysis_floats}

The reduction in selection overhead is particularly pronounced at higher retention budgets. At 30\% retention, FP-Prune's selection latency is 13.23, 2.47, and 12.51\,ms on LLaVA-NeXT, Qwen, and InternVL, respectively, yielding $4.83\times$, $5.83\times$, and $5.52\times$ speedups over each model's fastest pruning baseline. Retaining more tokens requires fewer deletions, reducing deletion \mbox{steps and selection overhead.}

% Keep this figure-heavy column at natural leading; put spare space below it.
\begingroup
\baselineskip=12pt\normalbaselineskip=\baselineskip
\microtypesetup{expansion=true}\hfuzz=0pt\hbadness=1000
FP-Prune also consistently reduces TTFT and delivers substantial end-to-end speedups over Dense. On LLaVA-NeXT, it reduces TTFT by 3.89\%--13.98\% relative to the fastest pruning baseline at each budget. Across the same budgets, TTFT speedups over Dense are $1.39\times$--$1.65\times$ on Qwen and $1.84\times$--$2.42\times$ on InternVL. Full-evaluation end-to-end speedups over Dense are $2.42\times$--$3.43\times$, $1.28\times$--$1.45\times$, and $1.72\times$--$2.13\times$ on LLaVA-NeXT, Qwen, and InternVL, respectively. Detailed efficiency measurements and comparisons are provided in Tables~A1 and~A2 \mbox{of the supplementary material.}\par\endgroup
\AddToHookNext{build/column/before}{\raggedbottom}

%% file: figures/analysis_floats.tex
% Float placement only; original captions preserved.
\begin{figure}[b]
  \centering
  \captionsetup{skip=5bp}
  \includegraphics[width=\columnwidth,trim=0 6bp 0 0,clip]{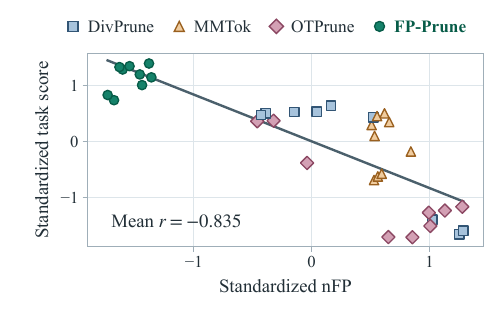}
  \caption{Association between normalized frame potential and downstream
  performance. Both metrics are averaged across eight datasets.
  Each point represents one method in one model--budget setting, and
  the line shows the linear fit \mbox{across the plotted points.}}
  \label{fig:fp_nfp_performance}
\end{figure}

\begin{figure*}[!t]
  \centering
  \captionsetup{skip=6pt}
  % Trim only empty space below the artwork; retain its horizontal scale.
  \includegraphics[width=\textwidth,trim=0 6bp 0 0,clip]{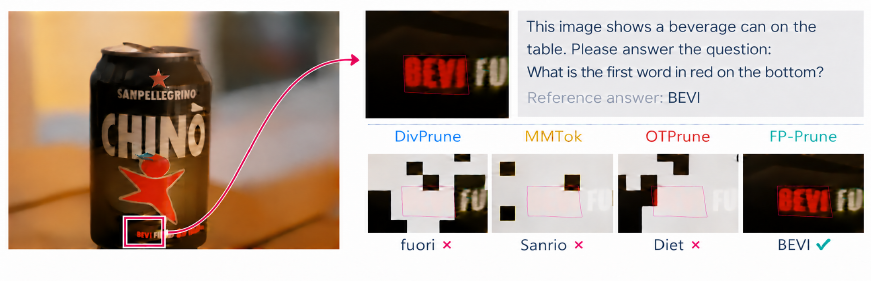}
  \caption{Retained regions and model predictions on a TextVQA example.
  The original image and a magnified view of the target word
  \emph{bevi} are shown alongside the question and reference answer.
  Each pruning method retains 333 of 3,328 visual tokens.
  The four method panels show the same crop, with outlines marking
  the target word. Faded areas lie outside the spatial footprints
  of the retained tokens. Model predictions are \mbox{shown below each panel.}}
  \label{fig:5}
  \vspace{12pt}
\end{figure*}

%% file: sections/5_analysis.tex
% Local heading spacing on the transition from the efficiency results.
\makeatletter
\let\FPsavedsection\cvprsection
\let\FPsavedsubsection\cvprsubsection
\patchcmd{\cvprsection}{-10pt}{-3pt}{}%
  {\PackageError{FP-Prune}{Cannot adjust analysis heading spacing}{Check cvpr.sty}}
\patchcmd{\cvprsection}{{7pt}}{{3pt}}{}%
  {\PackageError{FP-Prune}{Cannot adjust analysis heading spacing}{Check cvpr.sty}}
\patchcmd{\cvprsubsection}{{5pt}}{{3pt}}{}%
  {\PackageError{FP-Prune}{Cannot adjust analysis subheading spacing}{Check cvpr.sty}}
\makeatother
\section{Analysis}
\label{sec:fp_analysis}

\subsection{Frame Potential and Downstream Performance}
\makeatletter
\let\cvprsection\FPsavedsection
\let\cvprsubsection\FPsavedsubsection
\makeatother

We examine whether lower frame potential is associated with better
downstream performance at a fixed token retention budget. FP-Prune and
the three baselines are compared using normalized frame potential,
defined as $\mathrm{nFP}=\mathrm{FP}/K^2$, where $K$ is the number of
visual tokens retained per image, and the average task score Avg defined in
Sec.~\ref{sec:fp_setup}. Both metrics are averaged across the eight datasets.
We compute Pearson $r$ across the four methods separately for each of the
nine model--budget settings and report their mean in
Figure~\ref{fig:fp_nfp_performance}. The individual plots and correlations
are provided in the supplementary material.
Average nFP and average task score are negatively correlated in all nine
settings, with Pearson $r$ ranging from $-0.995$ to $-0.517$ and a mean
of $-0.835$. FP-Prune achieves the lowest nFP and highest average task
score in every setting. These results provide empirical support for the
pruning objective derived in Sec.~\ref{sec:fp_proxy}, where
minimizing frame potential maximizes a lower bound on retained information
\mbox{at a fixed token budget.}

% Let this subsection flow naturally in the arXiv page geometry.
\subsection{Coverage of Scene Text Regions}

FP-Prune's largest gain on TextVQA is on InternVL at 10\% token
retention, where it outperforms the strongest baseline by 8.48
percentage points (Sec.~\ref{sec:fp_results}). To investigate this
gain, we measure how much of each image region containing text
that matches a reference answer is covered by the spatial
footprints of the retained visual tokens.
As shown in Table~\ref{tab:4}, FP-Prune covers 55.92\% of these
regions on average and scores 61.43 on TextVQA,
compared with 40.13\% coverage and a score of 52.95 for DivPrune.
This suggests that greater coverage of image regions containing
the answer may contribute to the TextVQA gains.
Beyond answer-containing regions, FP-Prune also improves all-text
coverage from 40.67\% to 50.75\% and Hit@50 from 34.86\% to 51.40\%
over DivPrune. A hit requires coverage of at least half the text region.
These results show that its coverage advantage
extends to scene text more broadly, consistent with the gains on
\mbox{text-related tasks reported in Sec.~\ref{sec:fp_results}.}

\begingroup
\setlength{\intextsep}{7pt}
\begin{table}[H]
  \centering
  \input{tables/table4}
\end{table}
\endgroup

Figure~\ref{fig:5} illustrates this pattern on a question that
asks the model to read the first red word at the bottom of the
image. With the same budget of 333 visual tokens, FP-Prune fully
covers the target word \emph{bevi} and correctly answers BEVI,
while each baseline covers less than 7\% of the word and answers
incorrectly. This example highlights FP-Prune's advantage in
preserving fine-grained \mbox{visual information, particularly textual details.}

%% file: tables/table4.tex
\captionsetup{skip=4pt}
\caption{Scene text coverage on the analysis set and TextVQA scores
reproduced from Table~\ref{tab:fp_internvl}, for InternVL at 10\% token
retention. The coverage evaluation set and metrics are described \mbox{in the
supplementary material.}}
\label{tab:4}
\small
\setlength{\tabcolsep}{4.5pt}
\renewcommand{\arraystretch}{1.15}
\resizebox{\columnwidth}{!}{%
\begin{tabular}{lcccc}
\toprule
\multirow{2}{*}{\textbf{Method}} &
\multicolumn{3}{c}{\textbf{Scene text coverage (\%)}} &
\multirow{2}{*}{\textbf{TextVQA}} \\
\cmidrule(lr){2-4}
& \textbf{All text} & \textbf{Hit@50} & \textbf{Answer text} & \\
\midrule
DivPrune & 40.67 & 34.86 & 40.13 & 52.95 \\
MMTok & 14.86 & 7.68 & 15.97 & 34.18 \\
OTPrune & 23.17 & 18.87 & 21.78 & 21.64 \\
\rowcolor[gray]{0.93}
\textbf{FP-Prune} & \textbf{50.75} & \textbf{51.40} & \textbf{55.92} & \textbf{61.43} \\
\bottomrule
\end{tabular}%
}

%% file: sections/6_conclusion.tex
\section{Conclusion}
\label{conclusion}

This paper presents FP-Prune, a training-free visual token pruning method. We revisit visual token pruning from an information-gain perspective, using mutual information to characterize the incremental value of candidate tokens. We theoretically prove that frame potential can serve as an effective surrogate for mutual information. Building on this finding, FP-Prune iteratively removes the token whose deletion yields the largest decrease in frame potential to preserve as much complementary visual information as possible, while enabling efficient selection through incremental score updates. Experiments on three widely used vision-language models (VLMs) and eight datasets demonstrate that FP-Prune outperforms baseline methods in both accuracy and inference efficiency. Further analysis shows that, under the same token budget, lower frame potential consistently corresponds to better aggregate task performance. These results indicate that frame potential provides a theoretically grounded and computationally efficient selection criterion \mbox{for visual token pruning.}

\paragraph{Limitations.}
The current evaluation focuses on image understanding and does not establish the effectiveness of FP-Prune on video inputs. Extending the method to video understanding, where token selection must account for temporal dependencies across \mbox{frames, remains future work.}

%% file: sections/llm_usage.tex
\section*{LLM Usage Statement}
We used ChatGPT solely as a writing assistant for grammar checking and language polishing of the manuscript. All research ideas, experiments, analyses, and conclusions were conceived and conducted entirely by the authors. ChatGPT and other LLMs were not used for obtaining ideas, for technical content generation, \mbox{or for experimental design}.

%% file: supplementary_body.tex
% Merged appendix from the separately supplied, authoritative supplementary ZIP.
% Input after the main bibliography; the shared main document owns the preamble.
% Finish the references naturally, without balancing their final columns.
\par\clearpage
\appendix
% Continue PDF page numbers; distinguish appendix counters and hyperlink anchors.
\setcounter{table}{0}
\setcounter{figure}{0}
\setcounter{equation}{0}
\renewcommand{\thetable}{A\arabic{table}}
\renewcommand{\thefigure}{A\arabic{figure}}
\renewcommand{\theequation}{A\arabic{equation}}
\renewcommand{\theHtable}{appendix.\arabic{table}}
\renewcommand{\theHfigure}{appendix.\arabic{figure}}
\renewcommand{\theHequation}{appendix.\arabic{equation}}
\definecolor{FPHighlight}{RGB}{237,246,243}
\definecolor{FPGroup}{RGB}{242,244,246}

% Preserve the supplied full-width opening page without altering title macros
% or depending on the review-only lineno package.
\onecolumn
\begingroup
\centering
\Large
\textbf{FP-Prune: Rethinking Visual Token Pruning\\from an Information Gain Perspective}\\
\vspace{0.5em}Supplementary Material \\
\vspace{1.0em}\par
\begin{minipage}{\textwidth}
\normalfont\normalsize
\section{Complete efficiency measurements}
\input{tables/tableA1}
\end{minipage}
\par
\endgroup
\clearpage
\twocolumn[{
\begin{minipage}{\textwidth}
\normalfont\normalsize
\section*{A. Complete efficiency measurements (continued)}
\input{tables/tableA2}
\vspace{\dbltextfloatsep}
\end{minipage}
}]

% In the equation-heavy appendix, share the small vertical adjustment with
% display-math spacing rather than stretching words or paragraph gaps.
\begingroup
\renewcommand{\FPDisplaySpacing}{%
  \setlength{\abovedisplayskip}{8pt plus 1pt minus .5pt}%
  \setlength{\belowdisplayskip}{8pt plus 1pt minus .5pt}%
  \setlength{\abovedisplayshortskip}{5pt plus 1pt minus .5pt}%
  \setlength{\belowdisplayshortskip}{8pt plus 1pt minus .5pt}%
}
\FPDisplaySpacing
\input{sections/supp_information_derivations}
\par
\endgroup

\par\flushcolsend\clearpage
\twocolumn[{
\begin{minipage}{\textwidth}
\normalfont\normalsize
\section{Frame potential and downstream performance}
\noindent\begin{minipage}[t]{\dimexpr(\textwidth-\columnsep)/2\relax}
The main paper combines nine model--budget settings in a single plot using
within-setting standardization. Figure~\ref{fig:supp_fp_by_setting} presents
the separate comparisons on the original nFP and task-score scales. Each
panel compares FP-Prune with the three baselines at a fixed model \mbox{and token
retention budget.}
\end{minipage}\hfill
\begin{minipage}[t]{\dimexpr(\textwidth-\columnsep)/2\relax}
Both metrics are averaged across the
eight datasets, and each Pearson correlation is computed across four methods.
All nine settings exhibit a negative correlation, with coefficients ranging
from $-0.995$ to $-0.517$ and \mbox{an arithmetic mean of $-0.835$.}
\end{minipage}\par
\medskip

\begingroup
  \centering
  \captionsetup{type=figure,hypcap=false}
  \includegraphics[width=0.84\textwidth]{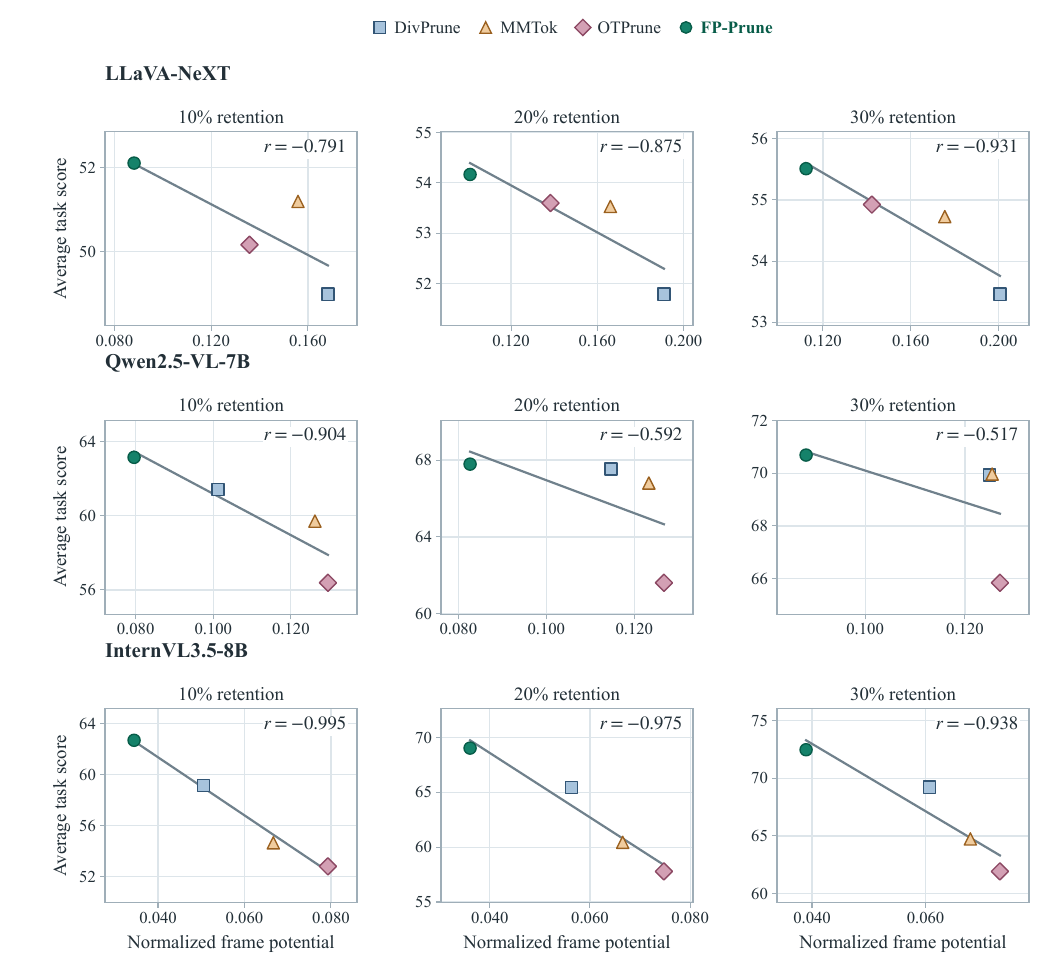}
  \caption{Normalized frame potential and downstream performance for each
  model and retention budget. Rows correspond to models and columns to
  retention budgets of 10\%, 20\%, and 30\%. Each point represents one pruning
  method, with both coordinates averaged across eight datasets.
  Axes show unstandardized values and are scaled independently in each panel.
  Lines show linear fits across the four methods, and annotations report
  \mbox{the corresponding Pearson correlations.}}
  \label{fig:supp_fp_by_setting}
\endgroup
\vspace{\dbltextfloatsep}
% Give the final two-column paragraph a shared heading and a common top baseline.
\section{Scene text coverage on TextVQA}
\end{minipage}
}]

We conduct the scene-text coverage analysis on the full TextVQA evaluation
split used for the results in Table~3 of the main paper. Coverage is measured
as the fraction of an annotated text region that falls within the spatial
footprints of retained visual tokens. We take the union of the footprints
from local tiles and the thumbnail before measuring coverage. Text region
coverage averages over text instances, and Hit@50 is the percentage of
instances with at least half of their area covered. Answer region coverage
averages coverage at the question level over questions for which a reference
answer can be matched to \mbox{text in the image.}

%% file: tables/tableA1.tex
\begingroup
\captionsetup{type=table,hypcap=false}
\caption{Token selection latency and TTFT on eight full evaluation sets. Latencies are in milliseconds and are averaged across benchmarks, weighted by their question counts. TTFT speedups are relative to Dense for the same model (Dense TTFT / pruned TTFT) and shown to two decimal places. Lower latency and higher speedup are better.}
\label{tab:A1}
\small
\setlength{\tabcolsep}{7pt}
\renewcommand{\arraystretch}{1.00}
\noindent\begin{tabular}{>{\raggedright\arraybackslash}p{\dimexpr.20\textwidth-2\tabcolsep\relax}>{\raggedright\arraybackslash}p{\dimexpr.24\textwidth-2\tabcolsep\relax}>{\raggedleft\arraybackslash}p{\dimexpr.19\textwidth-2\tabcolsep\relax}>{\raggedleft\arraybackslash}p{\dimexpr.18\textwidth-2\tabcolsep\relax}>{\raggedleft\arraybackslash}p{\dimexpr.19\textwidth-2\tabcolsep\relax}}
\toprule
\textbf{Retained budget} & \textbf{Method} & \textbf{Selection (ms) $\downarrow$} & \textbf{TTFT (ms) $\downarrow$} & \textbf{TTFT speedup $\uparrow$} \\
\midrule
\rowcolor{FPGroup}
\multicolumn{5}{l}{\strut\textbf{LLaVA-NeXT-Vicuna-7B}} \\
Full & Dense & --- & 827.97 & 1.00\(\times\) \\
\addlinespace[2pt]
10\% & DivPrune & 22.96 & 199.16 & 4.16\(\times\) \\
 & MMTok & 46.95 & 221.83 & 3.73\(\times\) \\
 & OTPrune & 44.57 & 218.47 & 3.79\(\times\) \\
\rowcolor{FPHighlight}
 & \textbf{FP-Prune} & \textbf{15.49} & \textbf{191.42} & \textbf{4.33\(\times\)} \\
\addlinespace[2pt]
20\% & DivPrune & 43.46 & 273.80 & 3.02\(\times\) \\
 & MMTok & 92.15 & 320.83 & 2.58\(\times\) \\
 & OTPrune & 83.57 & 310.81 & 2.66\(\times\) \\
\rowcolor{FPHighlight}
 & \textbf{FP-Prune} & \textbf{14.54} & \textbf{244.80} & \textbf{3.38\(\times\)} \\
\addlinespace[2pt]
30\% & DivPrune & 63.92 & 350.74 & 2.36\(\times\) \\
 & MMTok & 137.34 & 424.74 & 1.95\(\times\) \\
 & OTPrune & 122.89 & 408.19 & 2.03\(\times\) \\
\rowcolor{FPHighlight}
 & \textbf{FP-Prune} & \textbf{13.23} & \textbf{301.70} & \textbf{2.74\(\times\)} \\
\addlinespace[3pt]
\rowcolor{FPGroup}
\multicolumn{5}{l}{\strut\textbf{Qwen2.5-VL-7B}} \\
Full & Dense & --- & 281.28 & 1.00\(\times\) \\
\addlinespace[2pt]
10\% & DivPrune & 5.21 & 177.96 & 1.58\(\times\) \\
 & MMTok & 5.91 & 178.67 & 1.57\(\times\) \\
 & OTPrune & 8.99 & 181.83 & 1.55\(\times\) \\
\rowcolor{FPHighlight}
 & \textbf{FP-Prune} & \textbf{2.79} & \textbf{170.83} & \textbf{1.65\(\times\)} \\
\addlinespace[2pt]
20\% & DivPrune & 9.80 & 190.14 & 1.48\(\times\) \\
 & MMTok & 10.16 & 190.55 & 1.48\(\times\) \\
 & OTPrune & 17.03 & 197.56 & 1.42\(\times\) \\
\rowcolor{FPHighlight}
 & \textbf{FP-Prune} & \textbf{2.63} & \textbf{183.01} & \textbf{1.54\(\times\)} \\
\addlinespace[2pt]
30\% & DivPrune & 14.41 & 213.99 & 1.31\(\times\) \\
 & MMTok & 14.45 & 213.91 & 1.31\(\times\) \\
 & OTPrune & 25.11 & 224.61 & 1.25\(\times\) \\
\rowcolor{FPHighlight}
 & \textbf{FP-Prune} & \textbf{2.47} & \textbf{201.92} & \textbf{1.39\(\times\)} \\
\addlinespace[3pt]
\rowcolor{FPGroup}
\multicolumn{5}{l}{\strut\textbf{InternVL3.5-8B}} \\
Full & Dense & --- & 851.75 & 1.00\(\times\) \\
\addlinespace[2pt]
10\% & DivPrune & 25.35 & 368.03 & 2.31\(\times\) \\
 & MMTok & 55.04 & 397.76 & 2.14\(\times\) \\
 & OTPrune & 47.29 & 389.96 & 2.18\(\times\) \\
\rowcolor{FPHighlight}
 & \textbf{FP-Prune} & \textbf{14.84} & \textbf{352.69} & \textbf{2.42\(\times\)} \\
\addlinespace[2pt]
20\% & DivPrune & 47.13 & 446.81 & 1.91\(\times\) \\
 & MMTok & 107.01 & 506.59 & 1.68\(\times\) \\
 & OTPrune & 84.72 & 483.58 & 1.76\(\times\) \\
\rowcolor{FPHighlight}
 & \textbf{FP-Prune} & \textbf{13.66} & \textbf{415.78} & \textbf{2.05\(\times\)} \\
\addlinespace[2pt]
30\% & DivPrune & 69.08 & 516.23 & 1.65\(\times\) \\
 & MMTok & 158.80 & 606.78 & 1.40\(\times\) \\
 & OTPrune & 123.05 & 569.56 & 1.50\(\times\) \\
\rowcolor{FPHighlight}
 & \textbf{FP-Prune} & \textbf{12.51} & \textbf{462.03} & \textbf{1.84\(\times\)} \\
\bottomrule
\end{tabular}
\endgroup

%% file: tables/tableA2.tex
\begingroup
\captionsetup{type=table,hypcap=false}
\centering
\caption{End-to-end efficiency comparison between FP-Prune and Dense on eight full evaluation sets, with 29,698 questions per configuration. Time is reported as cumulative GPU time in GPU-seconds: the effective processing times on all GPUs are summed for each dataset and then aggregated across the eight datasets, excluding model loading, warm-up, and synchronization waits. Speedup is Dense cumulative GPU time divided by FP-Prune cumulative GPU time for the same model, shown to three decimal places. Lower cumulative GPU time and higher speedup are better.}
\label{tab:next_exact_timing}
\small
\setlength{\tabcolsep}{4pt}
\renewcommand{\arraystretch}{1.12}
\begin{tabular*}{\textwidth}{@{\extracolsep{\fill}}lrrrrrr@{}}
\toprule
\textbf{Method / budget} & \multicolumn{2}{c}{\textbf{LLaVA-NeXT-Vicuna-7B}} & \multicolumn{2}{c}{\textbf{Qwen2.5-VL-7B}} & \multicolumn{2}{c}{\textbf{InternVL3.5-8B}} \\
\cmidrule(lr){2-3}\cmidrule(lr){4-5}\cmidrule(lr){6-7}
 & \shortstack{\textbf{Time}\\\textbf{(GPU-seconds) $\downarrow$}} & \textbf{Speedup $\uparrow$} & \shortstack{\textbf{Time}\\\textbf{(GPU-seconds) $\downarrow$}} & \textbf{Speedup $\uparrow$} & \shortstack{\textbf{Time}\\\textbf{(GPU-seconds) $\downarrow$}} & \textbf{Speedup $\uparrow$} \\
\midrule
Dense / Full & 26,868.92 & 1.000\(\times\) & 9,996.37 & 1.000\(\times\) & 27,620.08 & 1.000\(\times\) \\
FP-Prune / 10\% & 7,843.07 & 3.426\(\times\) & 6,880.26 & 1.453\(\times\) & 12,955.04 & 2.132\(\times\) \\
FP-Prune / 20\% & 9,429.70 & 2.849\(\times\) & 7,343.19 & 1.361\(\times\) & 14,649.32 & 1.885\(\times\) \\
FP-Prune / 30\% & 11,122.13 & 2.416\(\times\) & 7,831.76 & 1.276\(\times\) & 16,061.11 & 1.720\(\times\) \\
\bottomrule
\end{tabular*}
\endgroup

%% file: sections/supp_information_derivations.tex
\section{Information retention under the Gaussian model}

\subsection{Mutual information of the retained set}
Let $G_S=U_SU_S^\top$, where the rows of $U_S$ are the normalized
features of the $K$ retained tokens. Under the Gaussian observation model
$T_S=U_SZ+\varepsilon_S$, the marginal covariance of $T_S$ is
$G_S+\sigma^2 I_K$. Conditioned on $Z$, the covariance is $\sigma^2 I_K$.
The Gaussian \mbox{entropy formula therefore gives}
\begin{equation*}
\begin{aligned}
I(Z;T_S)&=h(T_S)-h(T_S\mid Z)\\
&=\frac12\log\frac{\det(G_S+\sigma^2 I_K)}{\det(\sigma^2 I_K)}\\
&=\frac12\log\det\left(I_K+\frac{G_S}{\sigma^2}\right).
\end{aligned}
\end{equation*}

% \subsection*{B.2. Pairwise Information Redundancy}
% For $i\ne j$, the covariance of $T_i$ and $T_j$ is $u_i^\top u_j$, and
% each observation has variance $1+\sigma^2$. Their correlation is thus
% $\rho_{ij}=u_i^\top u_j/(1+\sigma^2)$. Applying the Gaussian entropy
% formula to the two observations gives
% \begin{equation*}
% I(T_i;T_j)=-\tfrac12\log(1-\rho_{ij}^2)
% =\tfrac12\rho_{ij}^2+O(\rho_{ij}^4).
% \end{equation*}
% When the correlations are small, summing the leading terms yields
% \begin{equation*}
% \mathcal{D}(S)\approx
% \frac{1}{2(1+\sigma^2)^2}\sum_{\substack{i,j\in S\\i<j}}(u_i^\top u_j)^2
% =\frac{FP(S)-K}{4(1+\sigma^2)^2}.
% \end{equation*}
% The final equality follows because the diagonal terms of $FP(S)$ sum to
% $K$ and each distinct pair contributes twice.

% section B2
\subsection{Derivation of the total-correlation--frame-potential relationship}
\label{app:tc_fp}

In this section, we provide the detailed derivation of
the relationship between mutual information, total
correlation, and Frame Potential under the Gaussian
observation model introduced \mbox{in the main text.}

\paragraph{Mutual information and total correlation.}

For a retained set $S$, the total correlation is defined as
\begin{equation}
\operatorname{TC}(T_S)
=
D_{\mathrm{KL}}\left(
p(T_S)\middle\|
\prod_{i\in S}p(T_i)
\right).
\end{equation}
We also define the conditional total correlation as
\begin{equation}
\begin{aligned}
&\operatorname{TC}(T_S\mid Z)\\
&=
\mathbb{E}_{Z}
\left[
D_{\mathrm{KL}}\left(
p(T_S\mid Z)\middle\|
\prod_{i\in S}p(T_i\mid Z)
\right)
\right].
\end{aligned}
\end{equation}

Using the definition of mutual information, we obtain
\begin{equation}
\begin{aligned}
&\sum_{i\in S}I(Z;T_i)-I(Z;T_S)\\
&=\operatorname{TC}(T_S)-\operatorname{TC}(T_S\mid Z).
\end{aligned}
\label{eq:tc_mi_identity}
\end{equation}

Under the Gaussian observation model, the token
representations are conditionally independent given $Z$,
since their observation noises \mbox{are mutually independent. Therefore,}
\begin{equation}
\operatorname{TC}(T_S\mid Z)=0.
\end{equation}

Furthermore, since $\|u_i\|_2=1$ and
$Z\sim\mathcal{N}(0,I_d)$, the marginal distribution
of each token representation is
\[
T_i\sim\mathcal{N}(0,1+\sigma^2),
\]
while its conditional variance given $Z$ is $\sigma^2$.
Consequently, the individual mutual information is
\begin{equation}
I(Z;T_i)
=
\frac{1}{2}\log\left(
1+\frac{1}{\sigma^2}
\right).
\end{equation}

Substituting these results into
Eq.~\eqref{eq:tc_mi_identity}, we obtain
\begin{equation}
I(Z;T_S)
=
\frac{K}{2}\log\left(
1+\frac{1}{\sigma^2}
\right)
-\operatorname{TC}(T_S),
\end{equation}
which establishes the exact relationship between
the retained mutual information and total correlation
under \mbox{the considered observation model.}

\paragraph{Gaussian expression of total correlation.}

Let $G_S=U_SU_S^\top$ denote the Gram matrix of the
normalized token features. According to the observation
model, the joint distribution of the retained token
representations is \mbox{Gaussian with covariance matrix}
\begin{equation}
\operatorname{Cov}(T_S)
=
G_S+\sigma^2 I_K.
\end{equation}

Since all token features are L2-normalized, each
marginal variance is $1+\sigma^2$. Using the Gaussian
KL divergence, the total correlation can therefore
be expressed in closed form as
\begin{equation}
\operatorname{TC}(T_S)
=
\frac{1}{2}\log
\frac{(1+\sigma^2)^K}
{\det(G_S+\sigma^2 I_K)}.
\label{eq:tc_logdet}
\end{equation}

\paragraph{Second-order approximation and frame potential.}

To establish the connection between total correlation
\mbox{and Frame Potential, let}
\begin{equation}
E_S=\frac{G_S-I_K}{1+\sigma^2}.
\end{equation}
Then, Eq.~\eqref{eq:tc_logdet} can be rewritten as
\begin{equation}
\operatorname{TC}(T_S)
=
-\frac{1}{2}\log\det(I_K+E_S).
\end{equation}

Since the diagonal entries of $G_S$ are all one,
we have $\operatorname{tr}(E_S)=0$.
Let $\lambda_1,\ldots,\lambda_K$ denote the eigenvalues
of $E_S$. Under the weak-correlation approximation,
a second-order Taylor expansion gives
\begin{equation}
\begin{aligned}
\operatorname{TC}(T_S)&=-\frac{1}{2}\sum_{k=1}^{K}\log(1+\lambda_k)\\
&\approx\frac{1}{4}\sum_{k=1}^{K}\lambda_k^2\\
&=\frac{1}{4}\operatorname{tr}(E_S^2).
\end{aligned}
\end{equation}

Substituting the definition of $E_S$ yields
\begin{equation}
\begin{aligned}
&\operatorname{tr}(E_S^2)\\
&=\frac{1}{(1+\sigma^2)^2}\operatorname{tr}\left((G_S-I_K)^2\right)\\
&=\frac{1}{(1+\sigma^2)^2}\left[\operatorname{tr}(G_S^2)-2\operatorname{tr}(G_S)+K\right].
\end{aligned}
\end{equation}

Since the normalized token features satisfy
$\|u_i\|_2=1$, we have
$\operatorname{tr}(G_S)=K$.
Moreover, the \mbox{Frame Potential is defined as}
\begin{equation}
FP(S)
=
\sum_{i,j\in S}(u_i^\top u_j)^2
=
\operatorname{tr}(G_S^2).
\end{equation}

Therefore, the trace can be written as
\begin{equation}
\operatorname{tr}(E_S^2)
=
\frac{FP(S)-K}{(1+\sigma^2)^2}.
\end{equation}

Combining the above results, we obtain
\begin{equation}
\operatorname{TC}(T_S)
\approx
\frac{FP(S)-K}{4(1+\sigma^2)^2}.
\end{equation}

This completes the derivation of the relationship
between total correlation and Frame Potential.

\subsection{Lower bound on retained information}
Let $\lambda_1,\ldots,\lambda_K$ be the nonnegative eigenvalues of $G_S$.
Since each feature has unit norm, $\sum_k\lambda_k=\operatorname{tr}(G_S)=K$.
Also, $\sum_k\lambda_k^2=\operatorname{tr}(G_S^2)=FP(S)$.
For $x\geq 0$, $\log(1+x)\geq x-x^2/2$ because the difference vanishes at
zero and has derivative $x^2/(1+x)\geq0$. Applying this bound to each
eigenvalue term gives
\begin{equation*}
\begin{aligned}
I(Z;T_S)&=\frac12\sum_{k=1}^K\log(1+\lambda_k/\sigma^2)\\
&\geq\frac{1}{2\sigma^2}\sum_{k=1}^K\lambda_k
-\frac{1}{4\sigma^4}\sum_{k=1}^K\lambda_k^2\\
&=\frac{K}{2\sigma^2}-\frac{FP(S)}{4\sigma^4}.
\end{aligned}
\end{equation*}
At fixed $K$ and $\sigma^2$, this lower \mbox{bound increases as $FP(S)$ decreases.}

%% file: main.bbl
\begin{thebibliography}{29}
\providecommand{\natexlab}[1]{#1}
\providecommand{\url}[1]{\texttt{#1}}
\expandafter\ifx\csname urlstyle\endcsname\relax
  \providecommand{\doi}[1]{doi: #1}\else
  \providecommand{\doi}{doi: \begingroup \urlstyle{rm}\Url}\fi

\bibitem[Alvar et~al.(2025)Alvar, Singh, Akbari, and Zhang]{ref8}
Saeed~Ranjbar Alvar, Gursimran Singh, Mohammad Akbari, and Yong Zhang.
\newblock {DivPrune}: Diversity-based visual token pruning for large multimodal models.
\newblock In \emph{Proceedings of the IEEE/CVF Conference on Computer Vision and Pattern Recognition}, pages 9392--9401, 2025.

\bibitem[Bai et~al.(2025)Bai, Chen, Liu, Wang, Ge, Song, Dang, Wang, Wang, Tang, Zhong, Zhu, Yang, Li, Wan, Wang, Ding, Fu, Xu, Ye, Zhang, Xie, Cheng, Zhang, Yang, Xu, and Lin]{ref3}
Shuai Bai, Keqin Chen, Xuejing Liu, Jialin Wang, Wenbin Ge, Sibo Song, Kai Dang, Peng Wang, Shijie Wang, Jun Tang, Humen Zhong, Yuanzhi Zhu, Mingkun Yang, Zhaohai Li, Jianqiang Wan, Pengfei Wang, Wei Ding, Zheren Fu, Yiheng Xu, Jiabo Ye, Xi Zhang, Tianbao Xie, Zesen Cheng, Hang Zhang, Zhibo Yang, Haiyang Xu, and Junyang Lin.
\newblock {Qwen2.5-VL} technical report.
\newblock \emph{arXiv preprint arXiv:2502.13923}, 2025.

\bibitem[Benedetto and Fickus(2003)]{benedetto2003finite}
John~J. Benedetto and Matthew Fickus.
\newblock Finite normalized tight frames.
\newblock \emph{Advances in Computational Mathematics}, 18\penalty0 (2--4):\penalty0 357--385, 2003.

\bibitem[Chen et~al.(2024{\natexlab{a}})Chen, Li, Dong, Zhang, Zang, Chen, Duan, Wang, Qiao, Lin, and Zhao]{chen2024mmstar}
Lin Chen, Jinsong Li, Xiaoyi Dong, Pan Zhang, Yuhang Zang, Zehui Chen, Haodong Duan, Jiaqi Wang, Yu Qiao, Dahua Lin, and Feng Zhao.
\newblock Are we on the right way for evaluating large vision-language models?
\newblock In \emph{Advances in Neural Information Processing Systems}, pages 27056--27087, 2024{\natexlab{a}}.

\bibitem[Chen et~al.(2024{\natexlab{b}})Chen, Zhao, Liu, Bai, Lin, Zhou, and Chang]{ref5}
Liang Chen, Haozhe Zhao, Tianyu Liu, Shuai Bai, Junyang Lin, Chang Zhou, and Baobao Chang.
\newblock An image is worth 1/2 tokens after layer 2: Plug-and-play inference acceleration for large vision-language models.
\newblock In \emph{Computer Vision -- ECCV 2024}, pages 19--35. Springer, 2024{\natexlab{b}}.

\bibitem[Chen et~al.(2026)Chen, Zhu, Li, Dong, Xiong, Wang, Qiu, Song, Wang, Tang, Wang, and Razi]{fp_chen2026otprune}
Xiwen Chen, Wenhui Zhu, Gen Li, Xuanzhao Dong, Yujian Xiong, Hao Wang, Peijie Qiu, Qingquan Song, Zhipeng Wang, Shao Tang, Yalin Wang, and Abolfazl Razi.
\newblock {OTPrune}: Distribution-aligned visual token pruning via optimal transport.
\newblock \emph{arXiv preprint arXiv:2602.20205}, 2026.

\bibitem[Cover and Thomas(2006)]{cover2006elements}
Thomas~M. Cover and Joy~A. Thomas.
\newblock \emph{Elements of Information Theory}.
\newblock Wiley, 2 edition, 2006.

\bibitem[Das et~al.(2025)Das, Bhatt, Kumari, Verma, and Bilmes]{Das_2025_CVPR}
Arnav~M. Das, Gantavya Bhatt, Lilly Kumari, Sahil Verma, and Jeff Bilmes.
\newblock {COBRA}: {COmBinatorial} retrieval augmentation for few-shot adaptation.
\newblock In \emph{Proceedings of the IEEE/CVF Conference on Computer Vision and Pattern Recognition}, pages 20534--20546, 2025.

\bibitem[Deng et~al.(2025)Deng, Li, Zhou, and He]{ref10}
Jinhong Deng, Wen Li, Joey~Tianyi Zhou, and Yang He.
\newblock {SCOPE}: Saliency-coverage oriented token pruning for efficient multimodel {LLMs}.
\newblock In \emph{Advances in Neural Information Processing Systems}, pages 161527--161552, 2025.

\bibitem[Dong et~al.(2026)Dong, Hu, Zhang, Yin, Fu, and Qian]{dong2026mmtok}
Sixun Dong, Juhua Hu, Mian Zhang, Ming Yin, Yanjie Fu, and Qi Qian.
\newblock {MMTok}: Multimodal coverage maximization for efficient inference of {VLMs}.
\newblock In \emph{International Conference on Learning Representations}, 2026.

\bibitem[Fu et~al.(2025)Fu, Chen, Shen, Qin, Zhang, Lin, Yang, Zheng, Li, Sun, Wu, Ji, Shan, and He]{fu2023mme}
Chaoyou Fu, Peixian Chen, Yunhang Shen, Yulei Qin, Mengdan Zhang, Xu Lin, Jinrui Yang, Xiawu Zheng, Ke Li, Xing Sun, Yunsheng Wu, Rongrong Ji, Caifeng Shan, and Ran He.
\newblock {MME}: A comprehensive evaluation benchmark for multimodal large language models.
\newblock In \emph{Advances in Neural Information Processing Systems}, 2025.

\bibitem[Guestrin et~al.(2005)Guestrin, Krause, and Singh]{guestrin2005near}
Carlos Guestrin, Andreas Krause, and Ajit~Paul Singh.
\newblock Near-optimal sensor placements in {Gaussian} processes.
\newblock In \emph{Proceedings of the 22nd International Conference on Machine Learning}, pages 265--272. ACM, 2005.

\bibitem[Hudson and Manning(2019)]{hudson2019gqa}
Drew~A. Hudson and Christopher~D. Manning.
\newblock {GQA}: A new dataset for real-world visual reasoning and compositional question answering.
\newblock In \emph{Proceedings of the IEEE/CVF Conference on Computer Vision and Pattern Recognition}, pages 6700--6709, 2019.

\bibitem[Li et~al.(2026)Li, Duan, Zhang, Ma, Xie, Carneiro, Yaqub, and Wang]{ref7}
Ao Li, Yuxiang Duan, Jinghui Zhang, Congbo Ma, Yutong Xie, Gustavo Carneiro, Mohammad Yaqub, and Hu Wang.
\newblock {TransPrune}: Token transition pruning for efficient large vision-language model.
\newblock In \emph{Proceedings of the IEEE/CVF Conference on Computer Vision and Pattern Recognition}, pages 39529--39538, 2026.

\bibitem[Li et~al.(2023)Li, Li, Savarese, and Hoi]{ref2}
Junnan Li, Dongxu Li, Silvio Savarese, and Steven Hoi.
\newblock {BLIP-2}: Bootstrapping language-image pre-training with frozen image encoders and large language models.
\newblock In \emph{Proceedings of the 40th International Conference on Machine Learning}, pages 19730--19742. PMLR, 2023.

\bibitem[Li et~al.(2025)Li, Zhan, Chen, Liu, Xiong, and Lu]{ref11}
Yangfu Li, Hongjian Zhan, Tianyi Chen, Qi Liu, Yu-Jie Xiong, and Yue Lu.
\newblock Why {$1+1<1$} in visual token pruning: Beyond naive integration via multi-objective balanced covering.
\newblock In \emph{Advances in Neural Information Processing Systems}, pages 71999--72039, 2025.

\bibitem[Liu et~al.(2023)Liu, Li, Wu, and Lee]{ref1}
Haotian Liu, Chunyuan Li, Qingyang Wu, and Yong~Jae Lee.
\newblock Visual instruction tuning.
\newblock In \emph{Advances in Neural Information Processing Systems}, pages 34892--34916, 2023.

\bibitem[Liu et~al.(2024{\natexlab{a}})Liu, Li, Li, Li, Zhang, Shen, and Lee]{liu2024llavanext}
Haotian Liu, Chunyuan Li, Yuheng Li, Bo Li, Yuanhan Zhang, Sheng Shen, and Yong~Jae Lee.
\newblock {LLaVA-NeXT}: Improved reasoning, {OCR}, and world knowledge.
\newblock Project blog, 2024{\natexlab{a}}.

\bibitem[Liu et~al.(2024{\natexlab{b}})Liu, Duan, Zhang, Li, Zhang, Zhao, Yuan, Wang, He, Liu, Chen, and Lin]{liu2024mmbench}
Yuan Liu, Haodong Duan, Yuanhan Zhang, Bo Li, Songyang Zhang, Wangbo Zhao, Yike Yuan, Jiaqi Wang, Conghui He, Ziwei Liu, Kai Chen, and Dahua Lin.
\newblock {MMBench}: Is your multi-modal model an all-around player?
\newblock In \emph{Computer Vision -- ECCV 2024}, pages 216--233. Springer, 2024{\natexlab{b}}.

\bibitem[Liu et~al.(2024{\natexlab{c}})Liu, Li, Huang, Yang, Yu, Li, Yin, Liu, Jin, and Bai]{liu2024ocrbench}
Yuliang Liu, Zhang Li, Mingxin Huang, Biao Yang, Wenwen Yu, Chunyuan Li, Xu-Cheng Yin, Cheng-Lin Liu, Lianwen Jin, and Xiang Bai.
\newblock {OCRBench}: on the hidden mystery of {OCR} in large multimodal models.
\newblock \emph{Science China Information Sciences}, 67:\penalty0 220102, 2024{\natexlab{c}}.

\bibitem[Lu et~al.(2022)Lu, Mishra, Xia, Qiu, Chang, Zhu, Tafjord, Clark, and Kalyan]{lu2022scienceqa}
Pan Lu, Swaroop Mishra, Tanglin Xia, Liang Qiu, Kai-Wei Chang, Song-Chun Zhu, Oyvind Tafjord, Peter Clark, and Ashwin Kalyan.
\newblock Learn to explain: Multimodal reasoning via thought chains for science question answering.
\newblock In \emph{Advances in Neural Information Processing Systems}, pages 2507--2521, 2022.

\bibitem[Ranieri et~al.(2014)Ranieri, Chebira, and Vetterli]{ranieri2014near}
Juri Ranieri, Amina Chebira, and Martin Vetterli.
\newblock Near-optimal sensor placement for linear inverse problems.
\newblock \emph{IEEE Transactions on Signal Processing}, 62\penalty0 (5):\penalty0 1135--1146, 2014.

\bibitem[Singh et~al.(2019)Singh, Natarajan, Shah, Jiang, Chen, Batra, Parikh, and Rohrbach]{singh2019textvqa}
Amanpreet Singh, Vivek Natarajan, Meet Shah, Yu Jiang, Xinlei Chen, Dhruv Batra, Devi Parikh, and Marcus Rohrbach.
\newblock Towards {VQA} models that can read.
\newblock In \emph{Proceedings of the IEEE/CVF Conference on Computer Vision and Pattern Recognition}, pages 8317--8326, 2019.

\bibitem[Wang et~al.(2025)Wang, Gao, Gu, Pu, Cui, Wei, Liu, Jing, Ye, Shao, Wang, Chen, Zhang, Yang, Wang, Wei, Yin, Li, Cui, Chen, Ding, Tian, Wu, Xie, Li, Yang, Duan, Wang, Hou, Hao, Zhang, Li, Zhao, Duan, Deng, Fu, He, Wang, He, Shi, He, Xiong, Lv, Wu, Shao, Zhang, Deng, Qi, Ge, Guo, Zhang, Zhang, Cao, Lin, Tang, Gao, Huang, Gu, Lyu, Tang, Wang, Lv, Ouyang, Wang, Dou, Zhu, Lu, Lin, Dai, Su, Zhou, Chen, Qiao, Wang, and Luo]{ref4}
Weiyun Wang, Zhangwei Gao, Lixin Gu, Hengjun Pu, Long Cui, Xingguang Wei, Zhaoyang Liu, Linglin Jing, Shenglong Ye, Jie Shao, Zhaokai Wang, Zhe Chen, Hongjie Zhang, Ganlin Yang, Haomin Wang, Qi Wei, Jinhui Yin, Wenhao Li, Erfei Cui, Guanzhou Chen, Zichen Ding, Changyao Tian, Zhenyu Wu, Jingjing Xie, Zehao Li, Bowen Yang, Yuchen Duan, Xuehui Wang, Zhi Hou, Haoran Hao, Tianyi Zhang, Songze Li, Xiangyu Zhao, Haodong Duan, Nianchen Deng, Bin Fu, Yinan He, Yi Wang, Conghui He, Botian Shi, Junjun He, Yingtong Xiong, Han Lv, Lijun Wu, Wenqi Shao, Kaipeng Zhang, Huipeng Deng, Biqing Qi, Jiaye Ge, Qipeng Guo, Wenwei Zhang, Songyang Zhang, Maosong Cao, Junyao Lin, Kexian Tang, Jianfei Gao, Haian Huang, Yuzhe Gu, Chengqi Lyu, Huanze Tang, Rui Wang, Haijun Lv, Wanli Ouyang, Limin Wang, Min Dou, Xizhou Zhu, Tong Lu, Dahua Lin, Jifeng Dai, Weijie Su, Bowen Zhou, Kai Chen, Yu Qiao, Wenhai Wang, and Gen Luo.
\newblock {InternVL3.5}: Advancing open-source multimodal models in versatility, reasoning, and efficiency.
\newblock \emph{arXiv preprint arXiv:2508.18265}, 2025.

\bibitem[Watanabe(1960)]{watanabe1960information}
Satosi Watanabe.
\newblock Information theoretical analysis of multivariate correlation.
\newblock \emph{IBM Journal of research and development}, 4\penalty0 (1):\penalty0 66--82, 1960.

\bibitem[Wen et~al.(2025)Wen, Gao, Wang, Zhang, Zhang, Li, He, and Zhang]{ref9}
Zichen Wen, Yifeng Gao, Shaobo Wang, Junyuan Zhang, Qintong Zhang, Weijia Li, Conghui He, and Linfeng Zhang.
\newblock Stop looking for ``important tokens'' in multimodal language models: Duplication matters more.
\newblock In \emph{Proceedings of the 2025 Conference on Empirical Methods in Natural Language Processing}, pages 9961--9980. Association for Computational Linguistics, 2025.

\bibitem[Yue et~al.(2024)Yue, Ni, Zhang, Zheng, Liu, Zhang, Stevens, Jiang, Ren, Sun, Wei, Yu, Yuan, Sun, Yin, Zheng, Yang, Liu, Huang, Sun, Su, and Chen]{yue2024mmmu}
Xiang Yue, Yuansheng Ni, Kai Zhang, Tianyu Zheng, Ruoqi Liu, Ge Zhang, Samuel Stevens, Dongfu Jiang, Weiming Ren, Yuxuan Sun, Cong Wei, Botao Yu, Ruibin Yuan, Renliang Sun, Ming Yin, Boyuan Zheng, Zhenzhu Yang, Yibo Liu, Wenhao Huang, Huan Sun, Yu Su, and Wenhu Chen.
\newblock {MMMU}: A massive multi-discipline multimodal understanding and reasoning benchmark for expert {AGI}.
\newblock In \emph{Proceedings of the IEEE/CVF Conference on Computer Vision and Pattern Recognition}, pages 9556--9567, 2024.

\bibitem[Zhang et~al.(2025)Zhang, Fan, Ma, Zheng, Huang, Cheng, Gudovskiy, Okuno, Nakata, Keutzer, and Zhang]{ref6}
Yuan Zhang, Chun-Kai Fan, Junpeng Ma, Wenzhao Zheng, Tao Huang, Kuan Cheng, Denis~A Gudovskiy, Tomoyuki Okuno, Yohei Nakata, Kurt Keutzer, and Shanghang Zhang.
\newblock {SparseVLM}: Visual token sparsification for efficient vision-language model inference.
\newblock In \emph{Proceedings of the 42nd International Conference on Machine Learning}, pages 74840--74857. PMLR, 2025.

\bibitem[Zhu et~al.(2026)Zhu, You, Jiang, Yang, Liu, and Yuan]{ref12}
Qingchan Zhu, Weihang You, Hanqi Jiang, Changdi Yang, Tianming Liu, and Geng Yuan.
\newblock Who speaks for the pruned? visual token pruning as coverage optimization.
\newblock In \emph{Proceedings of the 2026 Conference on Empirical Methods in Natural Language Processing}, pages 8018--8037. Association for Computational Linguistics, 2026.

\end{thebibliography}
